\documentclass{article}

\usepackage{microtype}
\usepackage{graphicx}
\usepackage{subfigure}
\usepackage{booktabs} 
\usepackage{xcolor}
\usepackage[utf8]{inputenc} 
\usepackage[T1]{fontenc}    
\usepackage{hyperref}       
\usepackage{url}            
\usepackage{amsfonts}       
\usepackage{nicefrac}       
\usepackage{microtype}      
\usepackage{algorithm}
\usepackage{algorithmic}
\usepackage{bbm}
\usepackage{amsmath}
\usepackage{amsthm}

\usepackage{hyperref}

\usepackage[accepted]{icml2021}
\newtheorem{thm}{Theorem}

\newtheorem{cor}{Corollary}
\newtheorem{rmk}{Remark}

\icmltitlerunning{Thompson Sampling under Bernoulli Rewards with Local Differential Privacy}

\begin{document}

\twocolumn[
\icmltitle{Thompson Sampling under Bernoulli Rewards with Local Differential Privacy}



\icmlsetsymbol{equal}{*}

\begin{icmlauthorlist}
\icmlauthor{Bo Jiang}{equal,to}
\icmlauthor{Tianchi Zhao}{equal,to}
\icmlauthor{Ming Li}{to}

\end{icmlauthorlist}

\icmlaffiliation{to}{Department of Electrical and Computer Engineering, University of Arizona, Arizona, USA}

\icmlcorrespondingauthor{Bo Jiang}{bjiang.email.arizona.edu}
\icmlcorrespondingauthor{Tianchi Zhao}{tzhao7.email.arizona.edu}
\icmlcorrespondingauthor{Ming Li}{lim.email.arizona.edu}

\icmlkeywords{Machine Learning, ICML}

\vskip 0.3in
]



\printAffiliationsAndNotice{\icmlEqualContribution} 

\begin{abstract}
This paper investigates the problem of regret minimization for multi-armed bandit
(MAB) problems with local differential privacy (LDP) guarantee. Given a fixed privacy budget $\epsilon$, we consider three privatizing mechanisms under Bernoulli scenario: linear, quadratic and exponential mechanisms. Under each mechanism, we derive stochastic regret bound for Thompson Sampling algorithm. Finally, we simulate to illustrate the convergence of different mechanisms under different privacy budgets.
\end{abstract}

\section{Introduction}
\label{sec:ins}

The multi-armed bandit (MAB) problem addresses the balancing of the trade-off between exploration and exploitation and has been widely applied in many real-world scenarios, from recommendation systems and information retrieval to healthcare and finance. In the settings of a MAB model, there are $N$ arms available to the agent, and each arm’s reward subjects to a particular distribution with an unknown mean. At each time step, the agent selects one arm. Then a reward is observed by the agent. The agent's ultimate goal is to gather as much cumulative reward as possible or minimize the total regret, i.e., designing a strategy that can explore different arms and exploit well-rewarded arm(s).

Nevertheless, personalized MAB implementations such as recommendation system is a double-edged
sword: the gained utility also comes with the risk of privacy violation. 
Comparing to the offline
learning models, online learning methods directly interact with sensitive user data, e.g., user clicks or purchasing history, and timely update the models to adjust their output, which makes privacy
an even more serious concern. For example, physicians want to test the effect of different treatments, and he collects patients' health conditions after a certain treatment. However, one's heart-beat data may compromise one's living routine, such as daily exercise time, sleeping time, etc. Another example is stock recommendation. The system (agent) periodically suggests different stocks (arms) to the user. After the suggestion, he wants to learn the feedback about how many shares (can be 0) the user has bought. However, directly revealing may leak the user’s buying power, personal preference, and what kind of risks he is hedging against.  In this paper, we leverage privacy protection in the MAB problem, i.e., the MAB problem where the observable reward at each time satisfies certain privacy constraints. 

Among all the privacy notions, Differential Privacy\cite{Dwork2006,Dwork2008} has been accepted as the \textit{de facto} standard for quantifying privacy leakage in the privacy community. The advantage of DP is that it provides a rigorous privacy guarantee against the worst-case adversaries, and is amenable for mechanism design. Recently, the local version of DP, local differential privacy (LDP), has gained significant attention. The server/aggregator who collects data is considered untrusted by the users, who perturb their data before sending it to the server. LDP based mechanisms have been successfully adopted by Google's RAPPOR \cite{Rappor} for collecting web browsing behavior, and Apple's MacOS to identify popular emojis and media preferences in Safari \cite{Cormode:2018:PSL:3183713.3197390, LPSApple}. LDP also enables a variety of privacy-preserving mechanisms for both discrete and continuous-valued data. Such as randomized response mechanism \cite{Tianhao}, Laplacian mechanism \cite{Dwork2006}, Gaussian mechanism \cite{8375673}.


Non-private MAB problems have been studied for decades, among which, either
frequentist methods like UCB (Upper Confidence Bound)  or Bayesian methods like Thompson Sampling \cite{pmlr-v23-agrawal12} have been shown to achieve optimal regret performance (up to constant factors). There is also a line of works related to the regret bound of MAB algorithm \cite{Finite}. A privacy-preserving MAB can be described as, in each round, a privatized/perturbed version of the reward(s) is (are) observable, and each perturbed reward satisfies certain privacy requirements. The earliest work that studied LDP bandits is \cite{pmlr-v83-gajane18a}, which proposed an LDP bandit algorithm that works for arms with Bernoulli rewards. In \cite{DBLP:journals/corr/abs-1905-12298}, for bandits with bounded rewards, a Laplace mechanism and a Bernoulli mechanism are proposed, and corresponding UCB algorithms are developed. The upper and lower bound are derived based on UCB algorithms. In \cite{10.1145/3383313.3412254}, the statistical regret bounds are derived under either DP or LDP for collaborative LinUCB algorithm in the context of collaborative MAB. However, seldom of these works try to derive theoretically regret bound for privacy-preserving Thompson Sampling algorithm. The challenge is that TS is a Bayesian approach that involves the posterior updating at the agent by observing the reward. However, under the privacy-preserving framework, the observable reward is noisy, and the posterior distribution is not fixed but depends on the concrete mechanism (noisy distribution). In this paper, we consider different noisy models providing LDP guarantees. Then under each mechanism, we derive the posterior distribution and bound the corresponding probabilities causing the sub-optimal selection. In this paper, for a given privacy budget for LDP, we derive upper regret bounds for the Thompson Sampling algorithm.


\begin{figure*}[t]
    \centering
    \includegraphics[width=0.6\textwidth]{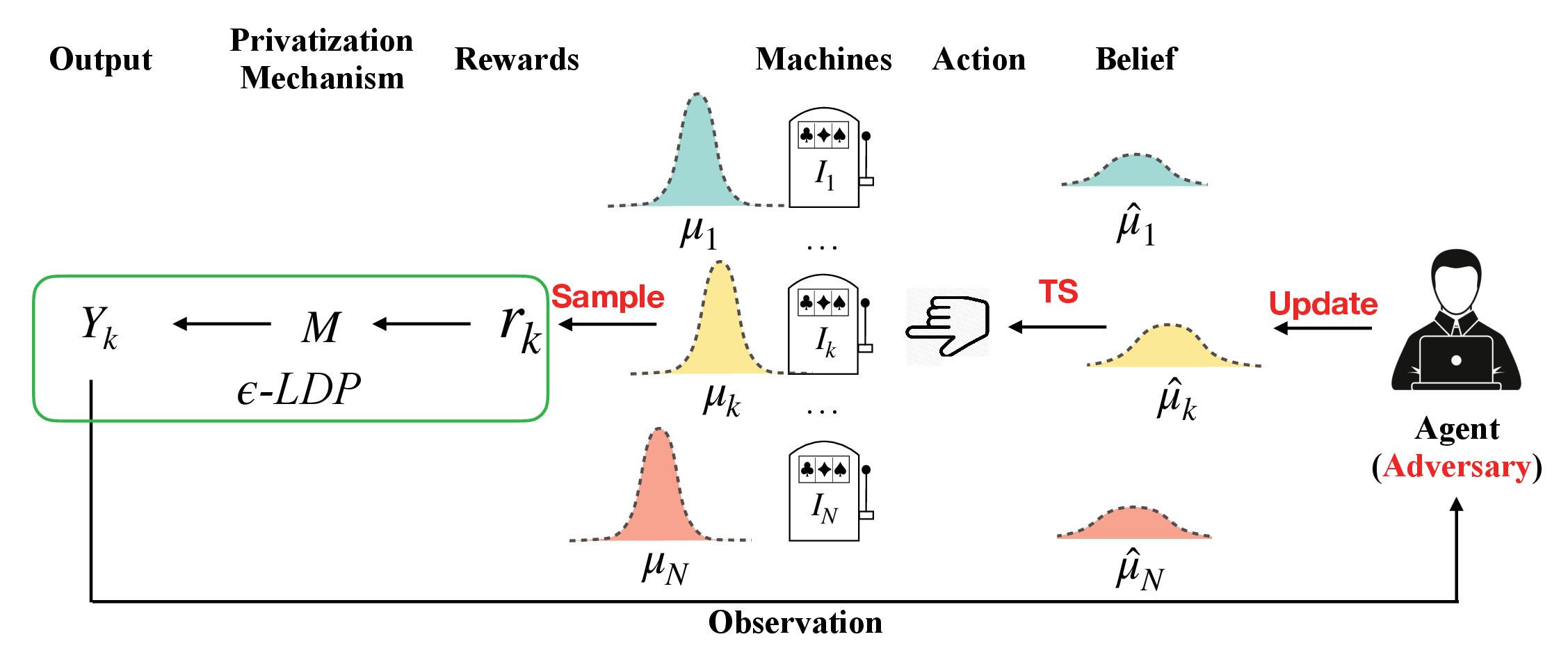}
    \caption{System model for privacy-preserving MAB instantaneous rewards are  privatized independently.}
    \label{fig:system_model}
    \vspace{-10pt}
\end{figure*}

The main contributions of this work are summarized as follows: 1). We propose different privacy-preserving MAB mechanisms under Thompson Sampling satisfying Local Differential Privacy; 2). We derive Cumulative Regret Bounds (CRB) for these mechanisms;
3). Simulate with synthetic data to support our analysis and compare the performance with $UCB$.



\section{Model Setup and Problem Formulation}
In this problem, we consider $N$ arms in the system, and each arm's reward $R\in\mathcal{R}$ follows a sub-Gaussian distribution with mean $\mu$. We use $\mu_i$ to denote the mean value of $R^i$, where $i\in\{1,2,...,N\}$ is the index of an arm. The agent, at each timestamp, selects one specific arm to play. The selected arm and the corresponding reward at time $k$ are denoted as $I_k\in\{1,2,...,N\}$ and $R_k$ respectively. 
Note that, in this problem, we assume the user (all the arms belong to one user) wants to cooperate with the agent to minimize the cumulative regret (in terms of TS algorithm, help the agent better learn the mean value). On the other hand, the user also wants each of his instantaneous reward's privacy to be protected. Therefore, the reward at each time is protected by a privacy-preserving mechanism $M$ ($M$ is assumed to be time-invariant). We define different kinds of privacy-preserving mechanisms later. Denote $Y_k\in\mathcal{Y}$ as the privatized version of $R_k$, which is also the output of $M$. The agent, after observing $Y_k$, can further update his belief state and the corresponding strategy for the next time.


In this work, we investigate the Thompson Sampling algorithm, which is a very classic algorithm for MAB problems. The algorithm can be summarized as follows: The agent has an estimated reward distribution for each arm (usually starting from a uniform distribution), denote $\hat{\mu}_i^k$ as the estimated mean reward for arm $i$ at time $k$. At each time, he randomly samples a reward for each arm from his estimated distribution and selects the arm that has the maximal sampled reward. After each round, by observing $Y_k$, he updates his belief state accordingly (the distribution of the reward of the arm that just played). 


To provide strict privacy protection to every instantaneous reward at each time, we let $M$ satisfy $\epsilon$-LDP, which can be defined as, for any $r,r'\in\mathcal{R}$, $y\in\mathcal{Y}$:
\begin{equation}
    Pr(Y_k=y|R_k=r)\le{e^{\epsilon}Pr(Y_k=y|R_k=r')},
\end{equation}
where $\epsilon$ is the privacy budget, the smaller $\epsilon$, the stronger privacy guarantee the mechanism provides. Note that the privacy-preserving mechanism is protecting the privacy of each instantaneous reward (sampled from a fixed distribution), not the distribution itself. On the contrary, TS algorithm requires an estimation of the posterior distribution, which tries to infer the data distribution by nature. From the privacy perspective, the privacy leakage of each instantaneous reward is guaranteed to be upper bounded by $\epsilon$, and the leakage of the distribution after observing $k$ samples from the same arm is upper bounded by $k\epsilon$ by the composability of LDP.
The system model is depicted in Fig.\ref{fig:system_model}.

\section{LDP-based Binomial Mechanisms}\label{Sec:algorithms}
In this section, we first introduce three LDP-based Binomial privacy-preserving mechanisms, including linear mechanism, quadratic mechanism, and exponential mechanism. Then, we discuss how to implement these mechanisms into the TS algorithm. In the following of this paper, we assume that the reward of each arm is bounded, and the first arm $I_1$ is the optimal arm ($\mu_1=\max_i{\mu_i}$). 

Bernoulli mechanism converts a bounded reward to a Bernoulli distributed one, i.e., $Y_k=Bernoulli(p(r))$, where $p(r)$ denotes the probability that $Y_k=1$ given the reward is $r$. {Algorithm \ref{alg:C} shows the detail of the algorithm for LDP based TS algorithm with Bernoulli mechanism.}
\begin{algorithm}
\caption{TS-LDP-B (LDP TS algorithm with Bernoulli mechanism)}  
\label{alg:C}  
\begin{algorithmic}[1]
\STATE {$S_i=0,F_i=0$}  
\FOR{ $t=1,2,...,$}
\STATE For each arm $i=1,\ldots, N$, sample $\theta_i(t)$ from the Beta($S_i+1$,$F_i+1$) distribution
\STATE Play arm $i(t):=\arg\max\theta_i(t)$ and observe $Y_t$. ($Y_t$ is the instance reward after performs Bernoulli privacy-preserving mechanisms)
\IF{$Y_t=1$}
\STATE $S_i=S_i+1$ ($S_i$ denotes the number of successes in the Bernoulli trial)
\ELSE
\STATE $F_i=F_i+1$  ($F_i$ denotes the number of failures in the Bernoulli trial)
\ENDIF
\ENDFOR
\end{algorithmic}  
\end{algorithm}

Next, in the following theorem, we present the sufficient condition to satisfy the $\epsilon$-LDP.

\begin{thm}\label{thm:ber}
For a bounded Bernoulli mechanism, to satisfy $\epsilon$-LDP, the following conditions must hold: (1) $p(0)\ge\frac{1}{e^{\epsilon}+1}$;
    (2) $p(1)\le\frac{e^{\epsilon}}{e^{\epsilon}+1}$; (3) $p(r)$ is monotonically increasing.
\end{thm}

In this paper, we consider three different probability functions under the Bernoulli mechanism. Linear probability function, Quadratic probability function, and Exponential probability function. Based on the sufficient conditions described in Theorem \ref{thm:ber}, $p(r)$ for each mechanism are stated in the following Corollary.

\begin{cor}
For linear probability function, $p(r)=\frac{1}{1+e^{\epsilon}}[(e^{\epsilon}-1)r+1]$;
For quadratic probability function, $p(r)=\frac{1}{1+e^{\epsilon}}[(e^{\epsilon}-1-b)r^2+br+1]$, where $b\in[0,2(e^{\epsilon}-1)]$; For exponential probability function, $p(r)=\frac{e^{\epsilon r}}{e^{\epsilon}+1}$.
\end{cor}

\begin{figure*}[t]
\centering
\subfigure[Regret bounds comparison for different $\epsilon$ TS algorithm under linear mechanism]{
\label{TSlinear}~~~~
\includegraphics[width=0.23\textwidth]{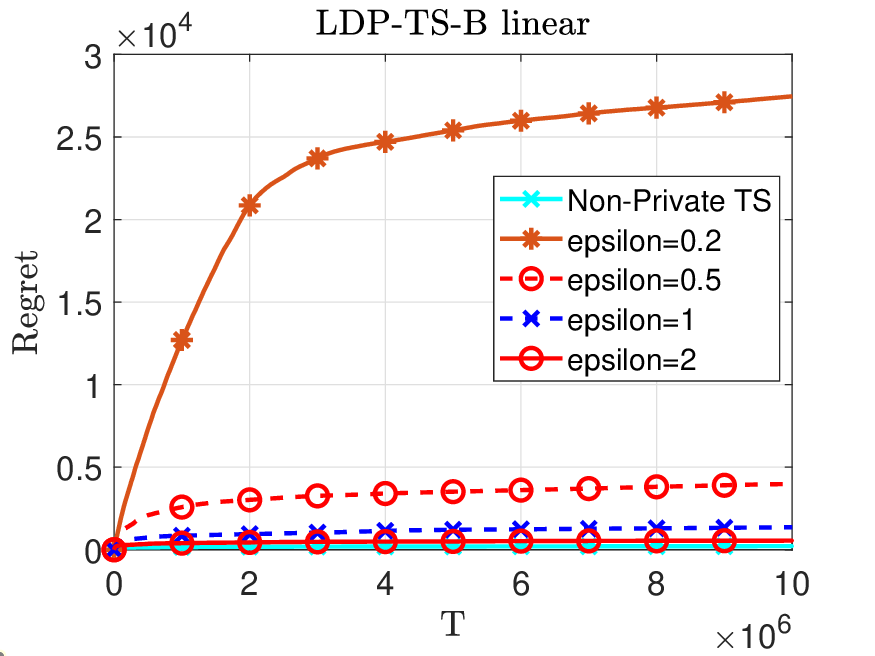}}
\subfigure[Regret bounds comparison for different $\epsilon$ TS algorithm under quadratic mechanism]{
\label{TSquad}~~~~
\includegraphics[width=0.23\textwidth]{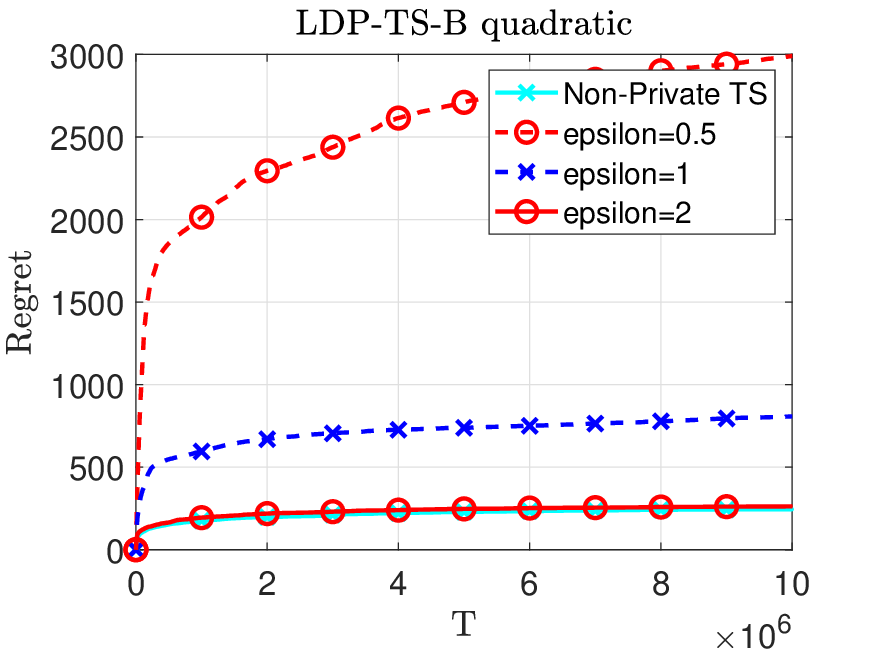}}
\subfigure[Regret bounds comparison for different $\epsilon$ TS algorithm under exponential mechanism]{
\label{TSexp}
\includegraphics[width=0.23\textwidth]{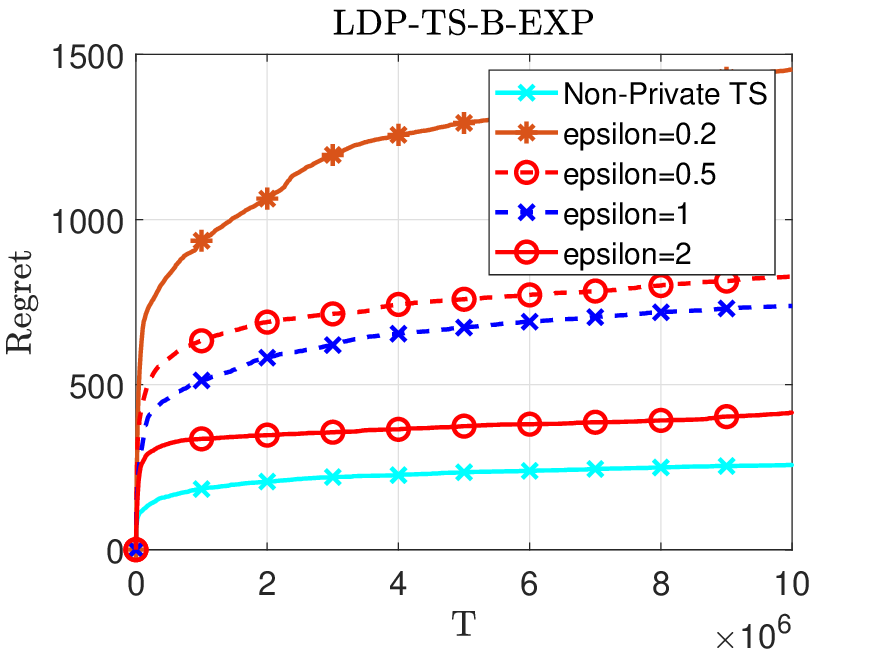}}
\subfigure[Regret bounds comparison for different $\epsilon$ UCB algorithm under linear mechanism]{
\label{UCBlinear}~~~~
\includegraphics[width=0.23\textwidth]{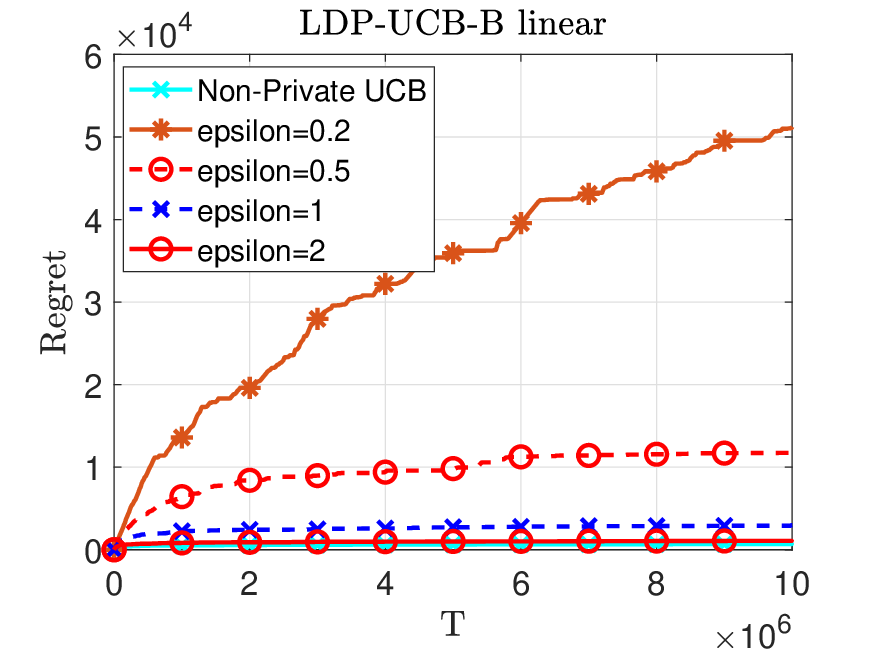}}
\subfigure[Regret bounds comparison for different $\epsilon$ UCB algorithm under quadratic mechanism]{
\label{UCBquad}~~~~
\includegraphics[width=0.23\textwidth]{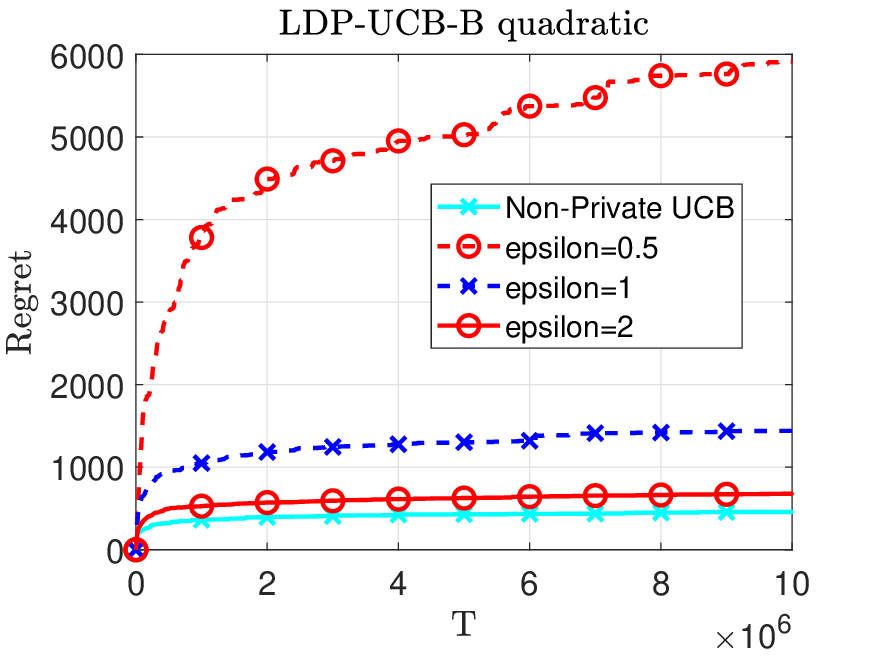}}
\subfigure[Regret bounds comparison for different $\epsilon$ UCB algorithm under exponential mechanism]{
\label{UCBexp}
\includegraphics[width=0.23\textwidth]{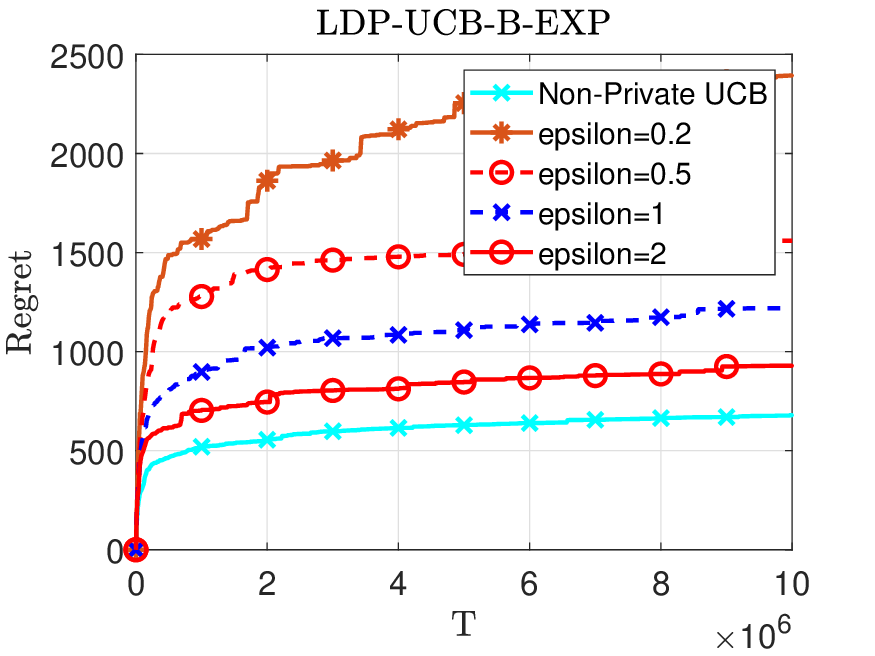}}
\label{SGSDUCB1111}

\vspace{-10pt}
\end{figure*}

\begin{rmk}
It is worth noting that the non-linear probability functions are preferable to the linear under certain circumstances. One scenario is that the mean rewards of different arms are very close to each other, the non-linear probability functions provide a better convergence rate compared to the linear model (it discriminates the optimal arm faster than the linear model). 
\end{rmk}

\section{Cumulative Regret Bounds}

Next, we derive CRB for TS-LDP-B, and we consider problem-dependent regret, where the regret at each time $t$ depends on the distance between the mean reward between arms $1$ and $i$; 
\subsection{Problem-dependent regret bound for linear probability function}
 The Problem-dependent regret bound for linear probability function is stated in the following theorem with proof provided in Appendix \ref{app_3}. 
\begin{thm}\label{thm3}
Given any non-zero $\Delta_i=\mu_1-\mu_i$, the cumulative regret for the linear probability function is upper bounded by ($\epsilon >0$):
\begin{equation}
    (1+\gamma)^2\left(\frac{e^{\epsilon}+1}{e^{\epsilon} -1}\right)^2\left\{\sum_{i\neq i^{*}}\frac{\log(T)}{2\Delta_{i}}+O\left(\frac{N}{2\Delta_{min}}\right)\right\}.
\end{equation}
Where $\Delta_{min}=\min_{i\in [N]} \Delta_i$, {$\gamma\in(0,1)$ is a threshold which helps to prove the regret bound.}
\end{thm}

\textbf{Proof outline for Theorem 2}: { The basic idea to prove the algorithm is similar to \cite{Agrawal2012FurtherThompson}. The difference of the proof for TS-LDP-B (linear probability function) is that we change the term in the denominator (from $d(\mu_i,\mu_1)$ to $\Delta_i$, which is based on the Pinsker's inequality). In this way, we can express {how the Bernoulli mechanism affects the regret bound.}}

\textbf{Remark: } {Note that, the regret bound of non-private TS is $ (1+\gamma)^2\left\{\sum_{i\neq i^{*}}\frac{log(T)}{2\Delta_{i}}+O(\frac{N}{2\Delta_{min}})\right\}$, wherein we change the denominator from $d(\mu_i,\mu_1)$ to $\Delta_i$ (loose version of the regret bound in \cite{Agrawal2012FurtherThompson}). Compared to non-private TS, the regret has the term $\left(\frac{e^{\epsilon}+1}{e^{\epsilon} -1}\right)^2$, which can be viewed as the cost for preserving privacy. When $\epsilon$ approaches infinity, this factor approaches 1, and the regret approaches that of the non-private version. When $\epsilon$ approaches zero, the regret becomes $O(T)$.}
\subsection{Problem-dependent regret bound for non-linear probability function}
  The Problem-dependent regret bound for the non-linear probability function is stated in the following Theorem with proof provided in Appendix \ref{app_5} and \ref{app_6}.
\begin{thm}\label{thm4}
Given any non-zero $\Delta_i=\mu_1-\mu_i$, the cumulative regret for the non-linear probability function is upper bounded by ($\epsilon >0$):
\begin{equation}
(1+\gamma)^2\sum_{i\neq 1}\frac{\log(T)+1}{2\Delta_{i,\epsilon}^2}\Delta_i +O(N),
\end{equation}
where $\Delta_{i,\epsilon}$ is the noisy difference between optimal arm and selected arm $i$.
\end{thm}

\textbf{Proof outline for Theorem 3}: { The proof structure of the non-linear probability function follows the similar idea of the linear probability function by changing the linear probability function to a non-linear probability function. }

\textbf{Remark:} {We can apply quadratic probability function and exponential probability function into (3). In quadratic probability function, $\Delta_{i,\epsilon}=\mu_{1,\epsilon,quad}-\mu_{i,\epsilon,quad}=\frac{\{(e^{\epsilon}-1-b)(\mu_1+\mu_i)+b\}(\mu_1-\mu_i)+(e^{\epsilon}-1-b)(\sigma_1^2-\sigma_i^2)}{e^{\epsilon} +1}$, $\mu_{i,\epsilon,quad}$ is the expected mean value of arm $i$ after performs quadratic probability function and $\sigma_i$ is arm $i$'s reward variance. we can see that it reduces to (2) when $e^{\epsilon}-1-b=0$. This result conforms to our expectation because the linear probability function is a special case for the quadratic probability function. In exponential probability function, $\Delta_{i,\epsilon,exp}=\mu_{1,\epsilon,exp}-\mu_{i,\epsilon,exp}=\frac{e^{\epsilon\mu_i}(e^{\epsilon\Delta_i}-1)+\tau_1(\epsilon)-\tau_i(\epsilon)}{e^{\epsilon} +1}$, $\mu_{i,\epsilon,exp}$ is the expected mean value of arm $i$ after performs exponential probability function and $\tau_i(\epsilon)$ is the  Jensen's gap between $e^{\epsilon\mu_i}$ and $E[e^{\epsilon r}]$. To make comparison with TS-LDP-B, we also provide UCB-LDP-B under non-linear probability function.  It is, }
 \begin{equation}
 \begin{aligned}
 \label{thm4ucb}
&R(T)\leq \sum_{i\neq 1}\left\{\frac{8\log(T)}{\Delta_{i,\epsilon}^2} +1+\frac{\pi^2}{3}\right\}\Delta_i.
\end{aligned}
 \end{equation} 
 Our analysis is based on Ren et al. \cite{ren2020multi} proof structure. However, the difference between our algorithm and their algorithm (LDP-UCB-B linear) is that we change the linear probability function to a non-linear probability function.
\section{NUMERICAL ANALYSIS}
In this section, we illustrate the numerical results of our algorithms. Due to computation limitations, we
only present the results for bandits with the Bernoulli mechanism. In the comparison, we compare LDP-TS-B studied in this paper to that of LDP-UCB-B \cite{ren2020multi}. We also include the performance of the non-private UCB and TS algorithm ($\epsilon=\infty$) as a baseline to see the cost for
preserving $\epsilon$-LDP.

The numerical results are illustrated in Fig. 2. In each experiment, we set the number of arms $N=20$. The optimal arm with a mean reward of $0.9$; five arms with $0.8$; another five arms with $0.7$; another five arms with $0.6$; the other four arms with $0.5$. We also let the rewards of arms follow different types of distributions: arms with mean rewards of $0.9$ or $0.6$ generate rewards from Bernoulli distributions; arms with mean rewards of $0.8$ generate rewards from Beta$(4, 1)$ distribution;
arms with mean rewards of $0.7$ generate rewards from $\{0.4, 1\}$ uniformly at random; and arms with
mean rewards of $0.5$ generate rewards from $[0, 1]$ uniformly at random. Curves in each figure are
averaged over 50 independent trials.

Fig. \ref{TSlinear} and Fig. \ref{UCBlinear} show the effect of different $\epsilon$ under linear model. We can see that the regret increases when $\epsilon$ decreases. This result is consistent with our theoretical results. Small $\epsilon$ has much more privacy, and the regret becomes large. Meanwhile, LDP-TS-B (linear) has lower regret than LDP-UCB-B (linear) under the same $\epsilon$.

Fig. \ref{TSquad} and Fig. \ref{UCBquad} show the effect of different $\epsilon$ under quadratic model. Same to the linear model, We can see that the regret increases when $\epsilon$ decreases, and LDP-TS-B (quadratic) has lower regret than LDP-UCB-B (quadratic) under the same $\epsilon$. 

Fig. \ref{TSexp} and Fig. \ref{UCBexp} show the effect of different $\epsilon$ under exponential model. Similar to the previous observation, We can see that the regret increases when $\epsilon$ decreases and LDP-TS-B (exponential) has lower regret than LDP-UCB-B (exponential) under the same $\epsilon$.
\section{Conclusion and Future Work}
In this paper, we studied the Thompson Sampling algorithm with local differential privacy guarantee. We consider three privatizing mechanisms under the Bernoulli rewards and 
proved a regret upper bounds for the Thompson Sampling algorithm. Numerical results also confirmed our theoretical results. For future work,  we are planning to derive a lower regret bound for general mechanisms.
\bibliographystyle{plainnat}
\bibliography{cited}
\begin{appendix}
\begin{small}
\section{Proof for Theorem \ref{thm3}}\label{app_3}
\begin{proof}
The $\epsilon$-LDP of LDP-TS-B follows from the $\epsilon$-DP of CTB stated in Lemma 5 \cite{ren2020multi}.

\textbf{Problem-dependent bound.} According to Lemma 5 \cite{ren2020multi}, for any arm $i$, the private response generated by CTB$(i,\epsilon)$ follows the Bernoulli$(\mu_{i,\epsilon})$ distribution, where \cite{ren2020multi}
\begin{equation}\label{eq:30}
    \begin{aligned}
    \mu_{i,\epsilon}:=\frac{1}{2}+\frac{(2\mu_i-1)(e^{\epsilon}-1)}{2(e^{\epsilon}+1)}
    \end{aligned}
\end{equation}
We define $\mu_{\epsilon}^{*}=\max_{i\in [N]}\mu_{i,\epsilon}$
\begin{equation}\label{eq:30}
    \begin{aligned}
    \Delta_{i,\epsilon}:=\mu_{\epsilon}^{*}-\mu_{i,\epsilon}=\frac{e^{\epsilon}-1}{e^{\epsilon} +1}\Delta_i
    \end{aligned}
\end{equation}
Next, we begin to derive regret of LDP-TS-B. The proof follows \cite{Agrawal2012FurtherThompson}. We fix a suboptimal arm $i\neq i^{*}$. Split according to $\hat{\mu}_{i,\epsilon}(t)$ and $p_{i,t}$
 \begin{equation}
 \begin{aligned}
 \label{eq:31}
&E[N_i(T)]=\sum_{t=1}^{T}Pr(I_t=i)\\
=&\sum_{t=1}^{T}\{Pr(I_t=i ,\hat{\mu}_{i,\epsilon}(t)> x_i)\nonumber\\+&Pr(I_t=i,\hat{\mu}_{i,\epsilon}(t)\leq x_i, p_{i,t}\geq y_i)\\+&Pr(I_t=i,\hat{\mu}_{i,\epsilon}(t)\leq x_i, p_{i,t}\leq y_i)\},
\end{aligned}
 \end{equation} 
 Where we choose $\mu_{i,\epsilon}<x_i<y_i<\mu_{\epsilon}^{*}$ as follows

\begin{equation}
 \begin{aligned}
 \label{eq:32}
&\mu_{i,\epsilon}<x_i<\mu_{\epsilon}^{*} \quad s.t.\quad d(x_i,\mu_{\epsilon}^{*})=\frac{d(\mu_{i,\epsilon},\mu_{\epsilon}^{*})}{1+\gamma}\\
&x_i<y_i<\mu_{\epsilon}^{*} \quad s.t.\quad d(x_i,y_i)=\frac{d(x_i,\mu_{\epsilon}^{*})}{1+\gamma}=\frac{d(\mu_{i,\epsilon},\mu_{\epsilon}^{*})}{(1+\gamma)^2}.
\end{aligned}
 \end{equation} 
 Where $0<\gamma <1$. Eq. \ref{eq:32} ensures that the relevant divergences are only a constant factor different from $d(\mu_{i,\epsilon},\mu_{\epsilon}^{*})$.
 
 For the first probability: denote $\tau_{i,k}$ as the k-th time when $I_t=i$, then
  \begin{equation}
 \begin{aligned}
 \label{eq:33}
&\sum_{t=1}^{T}Pr(I_t=i ,\hat{\mu}_{i,\epsilon}(t)> x_i)\\
\leq & 1+E\sum_{k=1}^{T-1}\mathbbm{1}\left[\hat{\mu}_{i,\epsilon}(t)> x_i\right]\nonumber\\\leq& 1+ \sum_{k=1}^{T-1}exp\left(-kd(x_i,\mu_{i,\epsilon})\right)\\
\leq & 1+\frac{1}{d(x_i,\mu_{i,\epsilon})}\stackrel{(a)} \leq 1+\frac{1}{d(x_i,y_i)} \nonumber\\\leq& 1+(1+\gamma)^2\frac{1}{d(\mu_{i,\epsilon},\mu_{\epsilon}^{*})}\\\stackrel{(b)}\leq & 1+(1+\gamma)^2\frac{1}{2(\mu_{i,\epsilon}-\mu_{\epsilon}^{*})^2},
\end{aligned}
 \end{equation} 
Inequality (a) is based on the fact that divergence is monotonically increasing function under $y_i>x_i$ when we fix $x_i$. Inequality (b) is based on the Pinsker's inequality. 
 
 For the second probability:

  \begin{equation}
 \begin{aligned}
 \label{eq:33}
&\sum_{t=1}^{T}Pr(I_t=i,\hat{\mu}_{i,\epsilon}(t)\leq x_i, p_{i,t}\geq y_i)\\
\leq & \sum_{t=1}^{T}Pr(I_t=i,p_{i,t}\geq y_i |\hat{\mu}_{i,\epsilon}(t)\leq x_i)\nonumber\\\leq& m+\sum_{t=\tau_{i,m}+1}^{T}Pr(I_t=i,p_{i,t}\geq y_i |\hat{\mu}_{i,\epsilon}(t)\leq x_i),
\end{aligned}
 \end{equation} 
 where we choose $m=\frac{log(T)}{d(x_i,y_i)}$ so that
   \begin{equation}
 \begin{aligned}
 \label{eq:33}
&\sum_{t=1}^{T}Pr(I_t=i,\hat{\mu}_{i,\epsilon}(t)\leq x_i, p_{i,t}\geq y_i)\\\leq& m+\sum_{t=\tau_{i,m}+1}^{T}Pr(I_t=i,p_{i,t}\geq y_i |\hat{\mu}_{i,\epsilon}(t)\leq x_i)\\\leq & m+E\left\{\sum_{t=\tau_{i,m}+1}^{T}exp(-N_i(t)d(x_i,y_i))\right\}\\\stackrel{(a)}\leq &\frac{\log(T)}{d(x_i,y_i)}+E\left\{\sum_{t=\tau_{i,m}+1}^{T}\frac{1}{T}\right\}\\\leq & \frac{\log(T)}{d(x_i,y_i)}+1\leq 1+ (1+\gamma)^2\frac{\log(T)}{d(\mu_{i,\epsilon},\mu_{\epsilon}^{*})}\\\leq& 1+(1+\gamma)^2\frac{\log(T)}{2(\mu_{i,\epsilon}-\mu_{\epsilon}^{*})^2}.
\end{aligned}
 \end{equation} 
 Inequality (a) is based on the fact that $exp(-N_i(t)d(x_i,y_i))\leq \frac{1}{T}$ when $t\geq \frac{log(T)}{d(x_i,y_i)}$.
 
 For the third probability, according to \cite{Agrawal2012FurtherThompson}, we have: 
 \begin{equation}\label{eq:34}
    \begin{aligned}
   \sum_{t=1}^{T}Pr(I_t=i,\hat{\mu}_{i,\epsilon}(t)\leq x_i, p_{i,t}\leq y_i)=O(1).
    \end{aligned}
\end{equation}
Combining, we get:
\begin{equation}
 \begin{aligned}
 \label{eq:35}
&E[N_i(T)]\leq (1+\gamma)^2\frac{\log(T)}{2(\mu_{i,\epsilon}-\mu_{\epsilon}^{*})^2}+(1+\gamma)^2\frac{1}{2(\mu_{i,\epsilon}-\mu_{\epsilon}^{*})^2}\\+&O(1),
\end{aligned}
 \end{equation} 
 The regret is 
 \begin{equation}
 \begin{aligned}
 \label{eq:36}
&R(T)\leq (1+\gamma)^2(\frac{e^{\epsilon}+1}{e^{\epsilon} -1})^2\left\{\sum_{i\neq i^{*}}\frac{\log(T)}{2\Delta_{i}}+O(\frac{N}{2\Delta_{min}})\right\},
\end{aligned}
 \end{equation} 
 \end{proof}
 \section{Proof for Theorem 3 under quadratic probability function}\label{app_5}
\begin{proof}
\textbf{Problem-dependent bound.}
Assumption, each arm $i$'s reward variance is $\sigma_i$.
\begin{equation}\label{eq:thm4mu}
    \begin{aligned}
    \mu_{i,\epsilon}:=\frac{(\mu_i^2+\sigma_i^2)(e^{\epsilon}-1-b)+b\mu_i+1}{e^{\epsilon}+1}
    \end{aligned}
\end{equation}
We define $\mu_{\epsilon}^{*}=\max_{i\in [N]}\mu_{i,\epsilon}$
\begin{equation}\label{eq:thm4regret2}
    \begin{aligned}
    &\Delta_{i,\epsilon}:=\mu_{\epsilon}^{*}-\mu_{i,\epsilon}=\\&\frac{\{(e^{\epsilon}-1-b)(\mu_1+\mu_i)+b\}(\mu_1-\mu_i)+(e^{\epsilon}-1-b)(\sigma_1^2-\sigma_i^2)}{e^{\epsilon} +1}
    \end{aligned}
\end{equation}
We put the Eq. \ref{eq:thm4regret2} value into Eq. \ref{eq:35} The regret is 
 \begin{equation}
 \begin{aligned}
 \label{thm4cum}
&R(T)\leq(1+\gamma)^2\sum_{i\neq i^{*}}\frac{\log(T)}{2\Delta_{i,\epsilon}^2}\Delta_i+\\&(1+\gamma)^2\sum_{i\neq i^{*}}\frac{1}{2\Delta_{i,\epsilon}^2}\Delta_i+O(N)\\=& (1+\gamma)^2\sum_{i\neq i^{*}}\frac{\log(T)+1}{2\left(\frac{\{(e^{\epsilon}-1-b)(\mu_1+\mu_i)+b\}(\mu_1-\mu_i)+(e^{\epsilon}-1-b)(\sigma_1^2-\sigma_i^2)}{e^{\epsilon} +1}\right)^2}\\&\Delta_i+O(N),
\end{aligned}
 \end{equation} 
 To prove the regret of UCB version, we use the expected time of sub-optimal arm in A.4 in \cite{ren2020multi}.
 \begin{equation}
 \begin{aligned}
 \label{eq:UCB_1}
&E[N_i(T)]\leq \frac{8\log(T)}{\Delta_{i,\epsilon}^2}+1+\frac{\pi^2}{3},
\end{aligned}
 \end{equation} 
 Then, the regret of UCB algorithm
 \begin{equation}
 \begin{aligned}
 \label{thm4cum}
&R(T)\leq \sum_{i\neq i^{*}}\{\frac{8\log(T)}{\left(\frac{\{(e^{\epsilon}-1-b)(\mu_1+\mu_i)+b\}(\mu_1-\mu_i)+(e^{\epsilon}-1-b)(\sigma_1^2-\sigma_i^2)}{e^{\epsilon} +1}\right)^2}\\& +1+\frac{\pi^2}{3}\}\Delta_i,
\end{aligned}
 \end{equation} 
{and if we assume all the arm i has same variance $\sigma$:}
\begin{equation}\label{eq:thm4regret}
    \begin{aligned}
    \Delta_{i,\epsilon}:=\mu_{\epsilon}^{*}-\mu_{i,\epsilon}=\frac{(e^{\epsilon}-1-b)(\mu_1+\mu_i)+b}{e^{\epsilon} +1}\Delta_i
    \end{aligned}
\end{equation}
 We put the Eq. \ref{eq:thm4regret} value into Eq. \ref{eq:35} The regret is 
 \begin{equation}
 \begin{aligned}
 \label{thm4cum}
&R(T)\leq(1+\gamma)^2\sum_{i\neq i^{*}}\frac{\log(T)}{2\Delta_{i,\epsilon}^2}\Delta_i+(1+\gamma)^2\sum_{i\neq i^{*}}\frac{1}{2\Delta_{i,\epsilon}^2}\Delta_i\\&+O(N)= (1+\gamma)^2\sum_{i\neq i^{*}}\frac{\log(T)+1}{2(\frac{(e^{\epsilon}-1-b)(\mu_{i^{*}}+\mu_i)+b}{e^{\epsilon} +1})^2\Delta_i^2}\Delta_i \\&+O(N) \leq (1+\gamma)^2(\frac{e^{\epsilon} +1}{(e^{\epsilon}-1-b)\mu_{i^{*}}+b})^2\sum_{i\neq i^{*}}\frac{\log(T)+1}{2\Delta_i} \\&+O(N),
\end{aligned}
 \end{equation} 
 To prove the regret of UCB version, we use the expected time of sub-optimal arm in A.4 in \cite{ren2020multi}.
 \begin{equation}
 \begin{aligned}
 \label{eq:UCB_1}
&E[N_i(T)]\leq \frac{8\log(T)}{\Delta_{i,\epsilon}^2}+1+\frac{\pi^2}{3},
\end{aligned}
 \end{equation} 
 Then, the regret of UCB algorithm
 \begin{equation}
 \begin{aligned}
 \label{thm4cum}
&R(T)\leq (\frac{e^{\epsilon} +1}{(e^{\epsilon}-1-b)\mu_{i^{*}}+b})^2\sum_{i\neq i^{*}}\frac{8\log(T)}{\Delta_i} +\\&\sum_{i\neq i^{*}}(1+\frac{\pi^2}{3})\Delta_i,
\end{aligned}
 \end{equation} 
 We assume all the arm's reward satisfies Bernoulli distribution, the variance of arm $i$ is $(1-\mu_i)\mu_i$, then the instance regret is 
 \begin{equation}\label{eq:thm4regret1}
    \begin{aligned}
    &\Delta_{i,\epsilon}:=\mu_{\epsilon}^{*}-\mu_{i,\epsilon}\\&=\frac{(e^{\epsilon}-1-b)(\mu_1^2-\mu_i^2+\sigma_1^2-\sigma_i^2)+b(\mu_1-\mu_i)}{e^{\epsilon} +1}\\=&\frac{(e^{\epsilon}-1-b)\left(\mu_1^2-\mu_i^2+\mu_1(1-\mu_1)-\mu_i(1-\mu_i)\right)+b(\mu_1-\mu_i)}{e^{\epsilon} +1}\\=&\frac{(e^{\epsilon}-1-b)(\mu_1-\mu_i)+b(\mu_1-\mu_i)}{e^{\epsilon} +1}=\frac{e^{\epsilon}-1}{e^{\epsilon} +1}(\mu_1-\mu_i)\\&=\frac{e^{\epsilon}-1}{e^{\epsilon} +1}\Delta_i
    \end{aligned}
\end{equation}
 \end{proof}
  \section{Proof for Theorem 3 under exponential probability function}\label{app_6}
\begin{proof}
\textbf{Problem-dependent bound.}
\begin{equation}\label{eq:thm5mu}
    \begin{aligned}
    \mu_{i,\epsilon}:=\frac{E[e^{\epsilon r}]}{e^{\epsilon}+1}=\frac{e^{\epsilon\mu_i}+\tau_i(\epsilon)}{e^{\epsilon}+1}
    \end{aligned}
\end{equation}
Where Eq. \ref{eq:thm5mu} is based on the Jensen's inequality ($e^{\epsilon\mu_i}\leq E[e^{\epsilon r}]$).  We define $\tau_i(\epsilon)$ as the  Jensen's gap between $e^{\epsilon\mu_i}$ and $E[e^{\epsilon r}]$. Then,
\begin{equation}\label{eq:thm5regret}
    \begin{aligned}
    &\Delta_{i,\epsilon}:=\mu_{\epsilon}^{*}-\mu_{i,\epsilon}=\frac{e^{\epsilon\mu_1}-e^{\epsilon\mu_i}+\tau_1(\epsilon)-\tau_i(\epsilon)}{e^{\epsilon} +1}\\&=\frac{e^{\epsilon(\Delta_i +\mu_i)}-e^{\epsilon\mu_i}+\tau_1(\epsilon)-\tau_i(\epsilon)}{e^{\epsilon} +1}\\= & \frac{e^{\epsilon\mu_i}(e^{\epsilon\Delta_i}-1)+\tau_1(\epsilon)-\tau_i(\epsilon)}{e^{\epsilon} +1}
    \end{aligned}
\end{equation}
 We put the Eq. \ref{eq:thm5regret} value into Eq. \ref{eq:35} The regret is 
 \begin{equation}
 \begin{aligned}
 \label{eq:thm5cum}
&R(T)\leq (1+\gamma)^2\sum_{i\neq i^{*}}\frac{\log(T)}{2\Delta_{i,\epsilon}^2}\Delta_i+(1+\gamma)^2\sum_{i\neq i^{*}}\frac{1}{2\Delta_{i,\epsilon}^2}\Delta_i\\&+O(N)= (1+\gamma)^2\sum_{i\neq i^{*}}\frac{\log(T)+1}{2(\frac{e^{\epsilon\mu_i}(e^{\epsilon\Delta_i}-1)+\tau_1(\epsilon)-\tau_i(\epsilon)}{e^{\epsilon} +1})^2}\Delta_i \\&+O(N) ,
\end{aligned}
 \end{equation} 
 Similarly, the regret of UCB algorithm
  \begin{equation}
 \begin{aligned}
 \label{eq:thm5cum1}
&R(T)\leq  \sum_{i\neq i^{*}}\left\{\frac{8\log(T)}{(\frac{e^{\epsilon\mu_i}(e^{\epsilon\Delta_i}-1)+\tau_1(\epsilon)-\tau_i(\epsilon)}{e^{\epsilon} +1})^2} +1+\frac{\pi^2}{3}\right\}\Delta_i ,
\end{aligned}
 \end{equation} 
 \end{proof}
 \end{small}
 \end{appendix}
\end{document}